\documentclass[11pt,a4paper]{article}
\usepackage{rotating}
\usepackage{acl2017}
\usepackage{times}
\usepackage{url}
\usepackage{amsmath}
\usepackage{amsthm}
\usepackage{amsfonts}
\usepackage{amssymb}
\usepackage{mathtools}
\usepackage{latexsym}
\usepackage{graphicx}
\usepackage{booktabs}
\usepackage[utf8]{inputenc}
\usepackage[small]{caption}
\usepackage{subcaption}
\usepackage{xspace}
\usepackage{enumitem}
\usepackage{float}
\usepackage{adjustbox}
\usepackage{appendix}

\usepackage{microtype}
\usepackage{arydshln}
\usepackage{booktabs}

\renewcommand{\cite}[2][]{\citep[#1]{#2}}   
\renewcommand{\newcite}[2][]{\citet[#1]{#2}}  

\usepackage{pifont}

\newcommand{\xmark}{\ding{55}}

\DeclareSymbolFont{extraup}{U}{zavm}{m}{n}
\DeclareMathSymbol{\varheart}{\mathalpha}{extraup}{86}
\DeclareMathSymbol{\vardiamond}{\mathalpha}{extraup}{87}

\usepackage[safe]{tipa}

\newcommand{\ignore}[1]{}

\def\aclpaperid{732} 

\aclfinalcopy

\usepackage[disable]{todonotes}
\newcommand{\note}[4][]{\todo[author=#2,color=#3,size=\scriptsize,fancyline,caption={},#1]{#4}} 
\renewcommand{\note}[4][]{}
   
\newcommand{\Jason}[2][]{}  
   
\newcommand{\Ryan}[2][]{}   
\newlength{\extramargin}
\makeatletter\if@todonotes@disabled\else
\usepackage{calc}
\usepackage[paperwidth=\paperwidth+\extramargin*2,marginparwidth=\marginparwidth+\extramargin,width=\textwidth,height=\textheight]{geometry} 
\fi\makeatother

\usepackage{bm}
\renewcommand{\vec}[1]{{\boldsymbol{\mathbf{#1}}}} 
\newcommand{\Real}{\mathbb{R}}
\newcommand{\e}{{\boldsymbol e}}
\newcommand{\f}{{\boldsymbol f}}

\newcommand{\eenergy}{\e}

\newcommand{\phone}[1]{\textipa{[#1]}}

\newcommand{\saveforCR}[1]{}

\usepackage{cleveref}

\crefname{section}{section}{sections}
\crefname{figure}{Figure}{Figs.}
\crefname{table}{Table}{Tables}
\crefname{algorithm}{Algorithm}{Algs.}
\crefname{equation}{equation}{eqs.}
\crefformat{section}{\S#2#1#3}  

\title{Probabilistic Typology: Deep Generative Models of Vowel Inventories}
\author{Ryan Cotterell \and Jason Eisner \\
Department of Computer Science \\
Johns Hopkins University \\
{\tt \{ryan.cotterell,eisner\}@jhu.edu}}

\begin{document}
\maketitle

\begin{abstract}
\vspace{-1pt}
  Linguistic typology studies the range of structures present in human
  language. The main goal of the field is to discover which sets of
  possible phenomena are universal, and which are merely frequent. For
  example, all languages have vowels, while most---but not
  all---languages have an \phone{u} sound. In this paper we present the first
  probabilistic treatment of a basic question in phonological
  typology: What makes a natural vowel inventory?  We introduce a
  series of deep stochastic point processes, and contrast them with
  previous computational, simulation-based approaches.  We provide a comprehensive suite of experiments on over 200 distinct languages.
\end{abstract}

\section{Introduction}\label{sec:intro}

Human languages exhibit a wide
range of phenomena, within some limits.  However, some
structures seem to occur or co-occur more frequently than others.
Linguistic typology attempts to describe the range of natural
variation and seeks to organize and quantify linguistic universals,
such as patterns of co-occurrence. Perhaps one of the simplest
typological questions comes from phonology: which vowels tend to occur
and co-occur within the phoneme inventories of different languages?
Drawing inspiration from the linguistic literature, we
propose models of the probability distribution from which the attested
vowel inventories have been drawn.

\begin{figure}[t]
  \vspace*{-10pt}
  \includegraphics[width=\columnwidth]{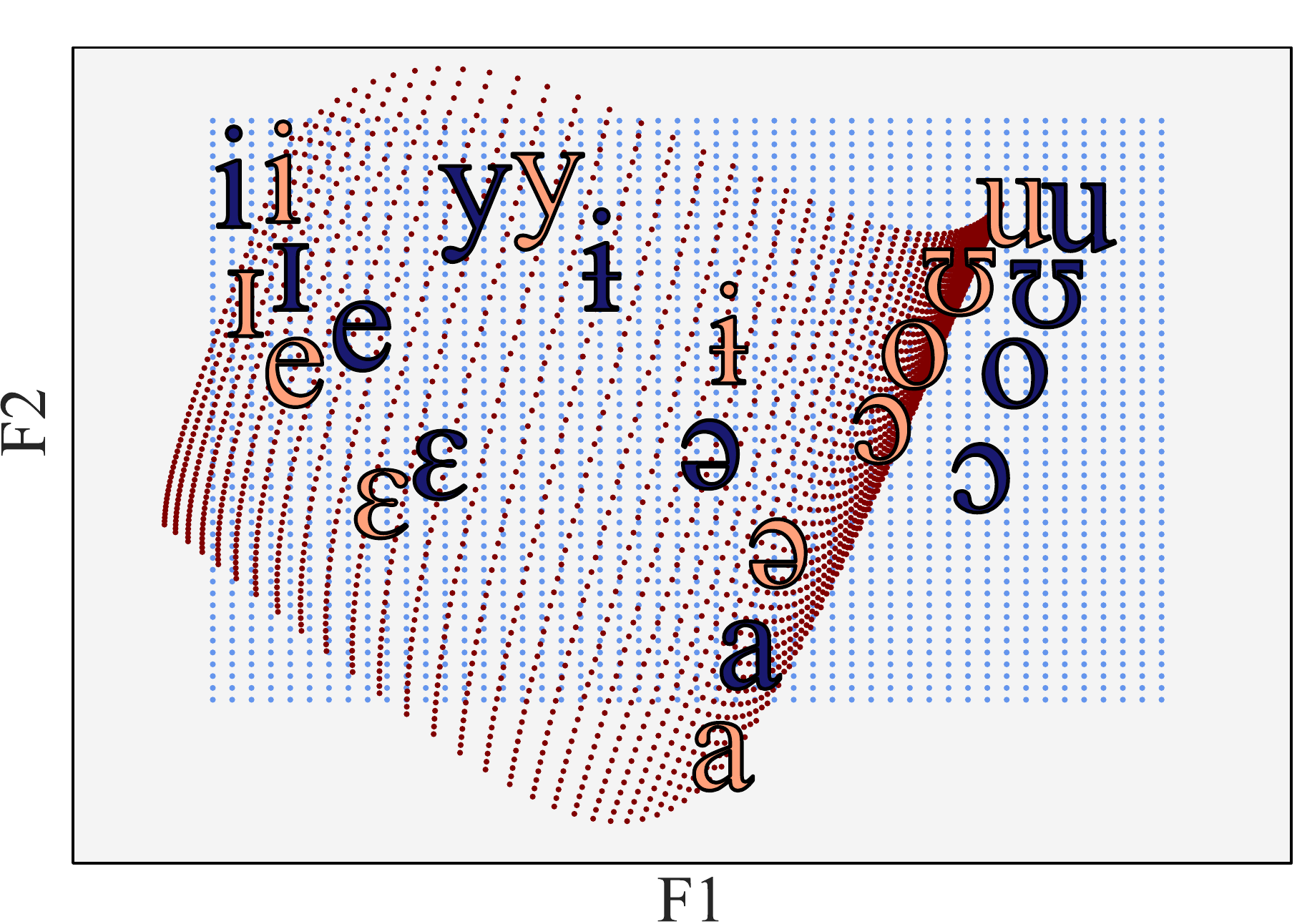}
  \caption{The transformed vowel space that is constructed within one of our deep generative models (see \cref{sec:results}).  A deep network nonlinearly maps the blue grid (``formant space'') to the red grid (``metric space''), with individual vowels mapped from blue to red position as shown.  Vowel pairs such as \phone{@}--\phone{O} that are brought close together are anti-correlated in the point process.  Other pairs such as \phone{y}--\phone{\textbari} are driven apart. For purposes of the visualization, we have transformed the red coordinate system to place red vowels near their blue positions---while preserving distances up to a constant factor (a ``Procrustes transformation'').}
  \label{fig:perceptual-space}
\vspace{-8pt}
\end{figure}

\begin{figure}
  \centering
  \includegraphics[width=.75\columnwidth]{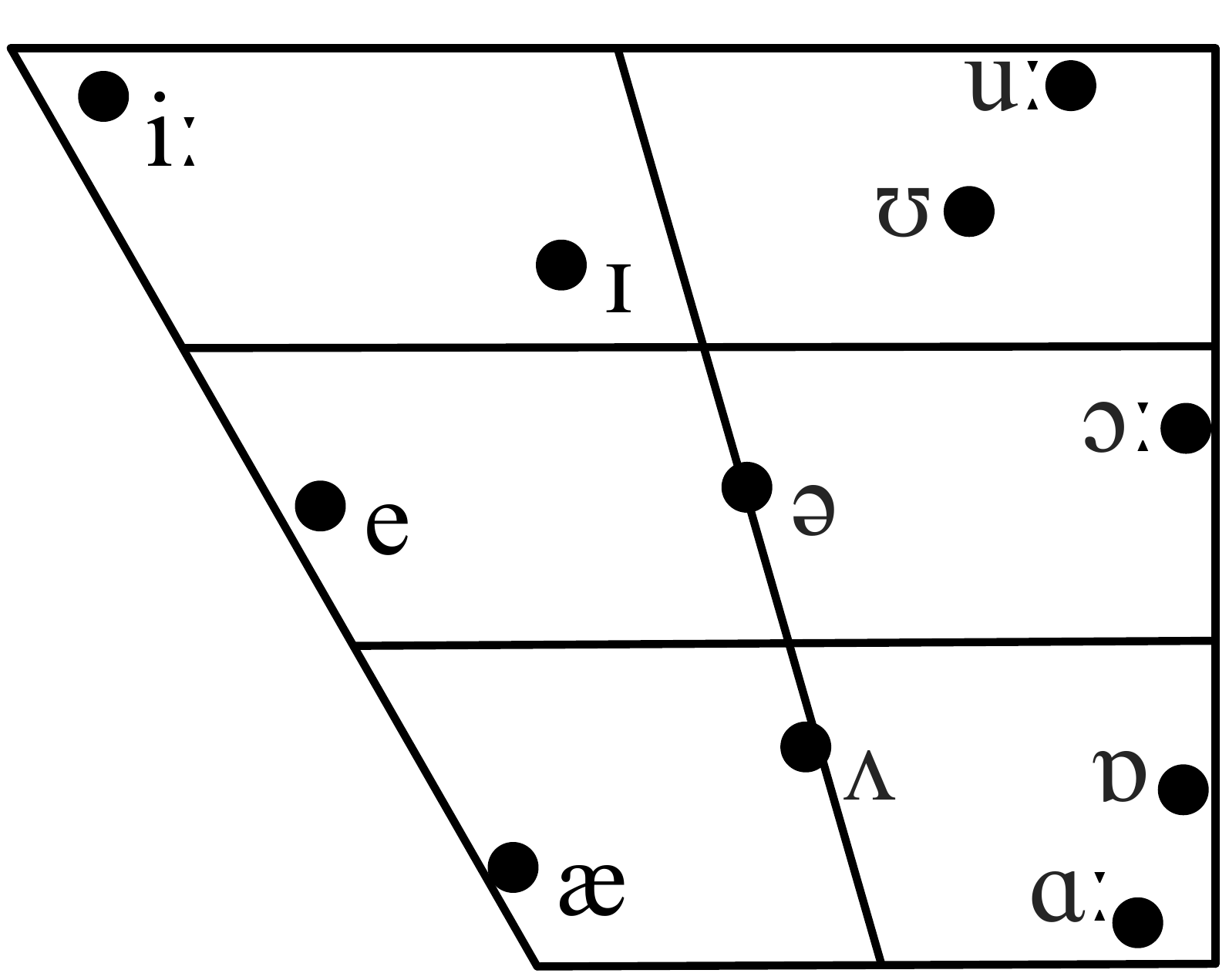}
  \caption{The standard vowel table in IPA for the RP accent of English. The $x$-axis indicates the front-back spectrum
    and the $y$-axis indicates the high-low distinction.}
  \label{fig:diagram}
\end{figure}

It is a typological universal that every language contains both vowels
and consonants \cite{velupillai2012introduction}.  But which vowels a
language contains is guided by softer constraints, in that certain
configurations are more widely attested than others. For instance, in
a typical phoneme inventory, there tend to be far fewer vowels than
consonants. Likewise, all languages contrast vowels based on height,
although which contrast is made is language-dependent
\cite{ladefoged1998sounds}. Moreover, while over 600 unique vowel
phonemes have been
attested cross-linguistically \cite{moran2014phoible}, certain regions
of acoustic space are used much more often than others, e.g., the
regions conventionally transcribed as \phone{a}, \phone{i}, and \phone{u}. Human
language also seems to prefer inventories where phonologically
distinct vowels are spread out in acoustic space (``dispersion'') 
so that they can be easily distinguished by a listener. We
depict the acoustic space for English in \cref{fig:diagram}.

In this work, we regard the proper goal of linguistic typology as the
construction of a universal prior distribution from which linguistic
systems are drawn.For vowel system typology, we propose
three formal probability models based on stochastic point processes.
We estimate the parameters of the model on one set of
languages and evaluate performance on a held-out set. We explore three
questions: (i) How well do the properties of our proposed probability
models line up experimentally with linguistic theory? (ii) How well
can our models predict held-out vowel systems? (iii) Do our models
benefit from a ``deep'' transformation from formant space to
metric space?

\section{Vowel Inventories and their Typology}\label{sec:vowel-inventories}
Vowel inventories are a simple entry point into the study of linguistic typology. Every spoken language chooses a discrete set of vowels, and the number of vowel phonemes ranges from 3 to 46, with a mean of 8.7 \cite{gordon}. Nevertheless, the empirical distribution over vowel inventories is remarkably peaked.  The majority of languages have 5--7 vowels, and there are only a handful of distinct 4-vowel systems attested despite many possibilities. Reigning linguistic theory \cite{becker2010acoustic} has proposed that vowel inventories are shaped by the principles discussed below.

\subsection{Acoustic Phonetics}

\begin{figure}
  \centering
  \includegraphics[width=1.\columnwidth]{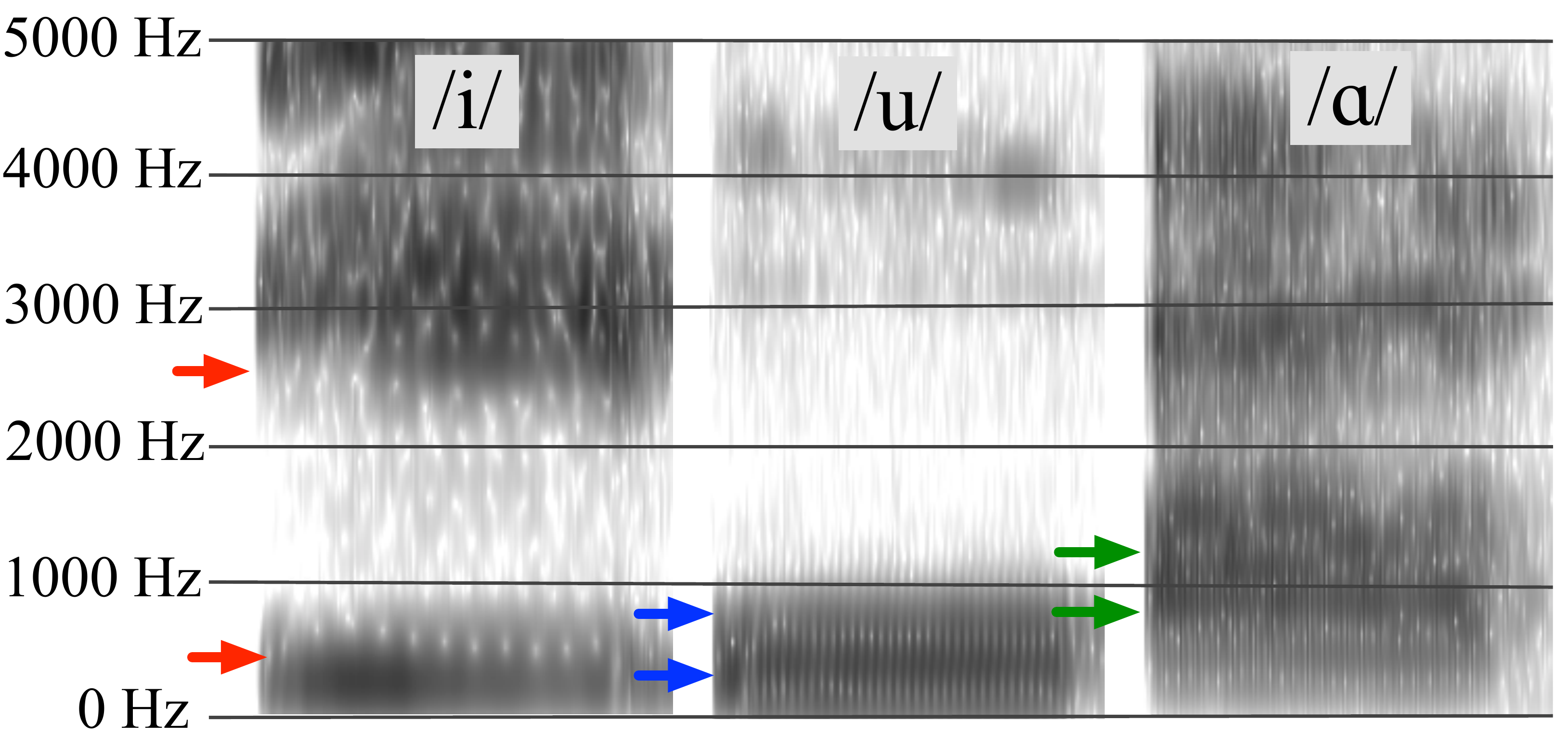}
  \caption{Example spectrogram of the three English vowels: \phone{i}, \phone{u} and \phone{A}. The $x$-axis is time and $y$-axis is frequency. The first two formants $F_1$ and $F_2$
    are marked in with colored arrows for each vowel. We used the Praat toolkit to generate the spectrogram
  and find the formants \cite{boersma2002praat}.}
\label{fig:spectrogram}
\end{figure}

One way to describe the sound of a vowel is through its acoustic energy at
different frequencies.  A spectrogram
(\cref{fig:spectrogram}) is a visualization of the energy at various
frequencies over time.  Consider the ``peak'' frequencies $F_0 < F_1 < F_2 < \ldots$
that have a greater energy than their neighboring frequencies.  
$F_0$ is called the fundamental frequency or pitch.  The other qualities of the vowel are largely determined by $F_1, F_2, \ldots$, which are known as formants \cite{ladefoged}.
In many languages, the first two formants
$F_1$ and $F_2$ contain enough information to identify a vowel:
\cref{fig:spectrogram} shows how these differ across three English vowels.
We consider each vowel listed in the International Phonetic
Alphabet (IPA) to be {\em cross-linguistically} characterized by some $(F_1,F_2)$ pair.

\subsection{Dispersion}\label{sec:dispersion}

The dispersion criterion \cite{liljencrants1972numerical,lindblom1986phonetic} states that the phonemes of a
language must be ``spread out'' so that they are easily
discriminated by a listener. 
A language seeks phonemes that are sufficiently ``distant'' from one another to
avoid confusion.
Distances between phonemes are defined in some latent ``metric space.''  We use this term rather than ``perceptual space'' because the confusability of two vowels may reflect not just their perceptual similarity, but also their common distortions by imprecise articulation or background noise.\footnote{We assume in this paper that the metric space is universal---although it would not be unreasonable to suppose that each language's vowel system has adapted to avoid confusion in the {\em specific} communicative environment of its speakers.}

\subsection{Focalization}\label{sec:focalization}

The dispersion criterion alone does not seem to capture the whole
story.  Certain vowels are simply more popular cross-linguistically.
A commonly accepted explanation is the {\em quantal theory of speech}
\cite{stevens1972quantal,stevens1989quanta}. The quantal theory states
that certain sounds are easier to articulate and to perceive than
others. These vowels may be characterized as those where $F_1$ and
$F_2$ have frequencies that are close to one another. On the
production side, these vowels are easier to pronounce since they allow
for greater articulatory imprecision.On the
perception side, they are more salient since the two spectral peaks
aggregate and act as one, larger peak to a certain degree.In general, languages will prefer these
vowels.

\subsection{Dispersion-Focalization Theory}\label{sec:dispersion-focalization}
The dispersion-focalization theory (DFT) combines both of the above notions. A good vowel system now consists of vowels that contrast with each other {\em and} are individually desirable \cite{schwartz1997dispersion}. This paper provides the first probabilistic treatment of DFT, and new evaluation metrics for future probabilistic and non-probabilistic treatments of vowel inventory typology.

\section{Point Process Models}\label{sec:point-processes}

Given a base set ${\cal V}$, a point process is a distribution over its subsets.\footnote{A point process is a specific kind of stochastic process, which is the technical term for a distribution over {\em functions}.  Under this view, 
  drawing some subset of ${\cal V}$ from the point process is regarded as drawing
  some {\em indicator function} on ${\cal V}$.
}
In this paper, we take ${\cal V}$ to be the set of all IPA
symbols corresponding to vowels.  Thus a draw from a point process is
a vowel inventory $V \subseteq {\cal V}$, and the point process itself
is a distribution over such inventories.  We
will consider three basic point process models for vowel systems: the 
Bernoulli Point Process, the Markov Point Process and the
Determinantal Point Process. In this section, we review the relevant
theory of point processes, highlighting aspects related to
\cref{sec:vowel-inventories}.

\subsection{Bernoulli Point Processes}
Taking ${\cal V} = \{v_1, \ldots, v_N\}$, 
a Bernoulli
point process (BPP) makes an {\em independent} decision
about whether to include each vowel in the subset. The
probability of a vowel system $V \subseteq {\cal V}$ is thus
\begin{equation}\label{eq:bpp}
  p(V) \propto \prod_{v_i \in V} \phi(v_i),
\end{equation}
where $\phi$ is a unary potential function, i.e., $\phi(v_i) \geq 0$.
Qualitatively, this means that $\phi(v_i)$ should be large if the $i^\text{th}$
vowel is good in the sense of \cref{sec:focalization}.
Marginal inference in a BPP is computationally trivial.  The probability that the inventory $V$ contains $v_i$ is $\phi(v_i)/(1+\phi(v_i))$, independent of the other vowels in $V$.
Since a BPP predicts each vowel independently, it only models focalization.  Thus, the model provides an appropriate baseline that will let us measure the importance of the dispersion principle---how far can we get with just focalization?  
A BPP may still tend to generate well-dispersed sets if it defines $\phi$ to be large only on certain vowels in $\cal V$ and these are well-dispersed (e.g., \phone{i}, \phone{u}, \phone{a}).  More precisely, it can define $\phi$ so that  $\phi(v_i) \phi(v_j)$ is small whenever $v_i, v_j$ are similar.\footnote{\label{fn:continuous}We point out that such a scheme would break down if we extended our work to cover fine-grained phonetic modeling of the vowel inventory.  In that setting, we ask not just whether the inventory includes /i/ but exactly which pronunciation of /i/ it contains.  In the limit, $\phi$ becomes a function over a {\em continuous} vowel space ${\cal V} = \Real^2$, turning the BPP into an inhomogeneous spatial Poisson process.  A continuous $\phi$ function implies that the model places similar probability on similar vowels.  Then if most vowel inventories contain some version of /i/, then many of them will contain several closely related variants of /i/ (independently chosen).  By contrast, the other methods in this paper do extend nicely to fine-grained phonetic modeling.}  But it cannot actively encourage dispersion: including $v_i$ does not lower the probability of also including $v_j$.  

\subsection{Markov Point Processes}
A Markov Point Process (MPP) \cite{opac-b1101505}---also known as a Boltzmann machine  \cite{ackley1985learning,pdp1986}---generalizes the BPP by adding pairwise interactions between vowels. The probability of
a vowel system $V \subseteq {\cal V}$ is now
\begin{equation}\label{eq:mpp}
  p(V) \propto \prod_{v_i \in V} \phi(v_i) \prod_{ v_i, v_j \in V} \psi(v_i, v_j),
\end{equation}
where each $\phi(v_i) \geq 0$ is, again, a unary potential that scores the
quality of the $i^\text{th}$ vowel, and each $\psi(v_i, v_j) \geq 0$ is a binary potential
that scores the combination of the $i^{\text{th}}$ and $j^{\text{th}}$
vowels. Roughly speaking, the potential $\psi(v_i, v_j)$ should be large if
the $i^{\text{th}}$ and $j^{\text{th}}$ vowel often co-occur.
Recall that under the principle of dispersion, the vowels that often co-occur
are easily distinguishable.  Thus, confusable vowel pairs
should tend to have potential $\psi(v_i, v_j) < 1$.

Unlike the BPP, the MPP can capture both focalization {\em and} dispersion. In this work, we will consider a fully connected MPP, i.e., there is a potential function for each pair of vowels in $\cal V$.  MPPs closely resemble Ising models \cite{ising1925beitrag}, but with the difference that Ising models are typically lattice-structured, rather than fully connected.

\paragraph{Inference in MPPs.}
Inference in fully connected MPPs, just as in general Markov Random
Fields (MRFs), is
intractable \cite{cooper1990computational} and we must rely on
approximation. In this work, we estimate any needed properties of the MPP
distribution by (approximately) drawing vowel inventories from it via Gibbs 
sampling
\cite{geman1984stochastic,Robert:2005:MCS:1051451}.
Gibbs sampling simulates a discrete-time Markov chain whose
stationary distribution is the desired MPP distribution.  
At each time step, for some random $v_i \in {\cal V}$, 
it stochastically decides whether to replace the current inventory $V$ with
$\bar{V}$, where $\bar{V}$ is a copy of $V$ with $v_i$ added (if $v_i
\notin V$) or removed (if $v_i \in V$).  The probability of replacement is
$\frac{p(\bar{V})}{p(V)+p(\bar{V})}$.  

\subsection{Determinantal Point Processes}
A determinantal point process (DPP) \cite{macchi1975coincidence} provides an elegant alternative to an MPP, and one that is directly suited to modeling both focalization and dispersion.  Inference
requires only a few matrix computations and runs tractably in $O(|{\cal V}|^3)$
time, even though the model may encode a rich set of multi-way interactions. 
We focus on the $L$-ensemble parameterization of the DPP, due to
\newcite{borodin2005eynard}.\footnote{Most DPPs are
  $L$-ensembles \cite{MAL-044}.} This type of DPP
  defines the probability of an inventory $V \subseteq \cal V$ as
\begin{equation}\label{eq:dpp}
  p(V) \propto \det L_{V},
\end{equation} 
where $L \in \Real^{N \times N}$ 
(for $N=|{\cal V}|$) is a symmetric
positive semidefinite matrix, 
and $L_V$ refers to the
submatrix of $L$ with only those rows and columns corresponding to
those elements in the subset $V$.

Although MAP inference remains NP-hard in DPPs 
(just as in MPPs), marginal inference becomes tractable.  We may compute the normalizing constant in closed form as follows:
\begin{equation}
  \sum_{V \in 2^{\cal V}} \det L_{V} = \det\left(L + I \right).
\end{equation}

How does a DPP ensure focalization and dispersion?  $L$ is positive semidefinite iff it can be written as $E^\top E$ for some matrix $E \in \Real^{N \times N}$.  It is possible to express $p(V)$ in terms of the column vectors of $E$, which we call $\vec{e}_1, \ldots, \vec{e}_N$:
\begin{itemize}[noitemsep]
\item For inventories of size 2, $p(\{v_i,v_j\}) \propto (\phi(v_i) \phi(v_j) \sin \theta)^2$, where $\phi(v_i), \phi(v_j)$ represent the {\em quality} of vowels $v_i, v_j$ (as in the BPP) while $\sin \theta \in [0,1]$ represents their {\em dissimilarity}.  More precisely, $\phi(v_i), \phi(v_j)$ are the lengths of vectors $\vec{e}_i, \vec{e}_j$ while $\theta$ is the angle between them.
Thus, we should choose the columns of $E$ so that focal vowels get {\em long} vectors and similar vowels get vectors {\em of similar direction}.
\item Generalizing beyond inventories of size 2, $p(V)$ is proportional to the square of the volume of the parallelepiped whose sides are given by $\{\vec{e}_i: v_i \in V\}$.  This volume can be regarded as $\prod_{v_i \in V} \phi(v_i)$ times a term that ranges from 1 for an orthogonal set of vowels to 0 for a linearly dependent set of vowels.
  \item The events $v_i \in V$ and $v_j \in V$ are anti-correlated (when not independent).  That is, while both vowels may individually have high probabilities (focalization), having either one in the inventory lowers the probability of the other (dispersion).
\end{itemize}

\section{Dataset}

At this point it is helpful to introduce the empirical dataset we will
model.  For each of 223 languages,\footnote{Becker-Kristal lists some languages multiple times with different measurements.  When a language had multiple listings, we selected one randomly for our experiments.} \newcite{becker2010acoustic} provides the vowel inventory as a 
set of IPA symbols, listing the first 5 formants for each vowel
(or fewer when not available in the original source).
Some corpus statistics are shown in
\cref{fig:vowel-frequency,fig:vowel-inventories}.\footnote{\label{fn:ipa}Caveat: The
  corpus is a curation of information from various phonetics papers
  into a common electronic format.  No standard procedure was followed
  across all languages: it was up to individual phoneticists to
  determine the size of each vowel inventory, the choice of IPA
  symbols to describe it, and the procedure for measuring the
  formants.  Moreover, it is an idealization to provide a single
  vector of formants for each vowel {\em type} in the language.  In real
  speech, different {\em tokens} of the same vowel are pronounced
  differently, because of coarticulation with the vowel context,
  allophony, interspeaker variation, and stochastic intraspeaker
  variation.  Even within a token, the formants change during the
  duration of the vowel.  Thus, one might do better to represent a
  vowel's pronunciation not by a formant vector, but by a conditional
  probability distribution over its formant trajectories given its
  context, or by a parameter vector that characterizes such a
  conditional distribution.  This setting would require richer data
than we present here.}
For the present paper, we take $\cal V$ to be the set of all 53 IPA
symbols that appear in the corpus.  We treat these IPA labels as
meaningful, in that we consider two vowels in different languages to
be the {\em same} vowel in $\cal V$ if (for example) they are both
annotated as \phone{O}.  We characterize that vowel by its {\em
  average} formant vector across all languages in the corpus that
contain the vowel: e.g., $(F_1, F_2, \ldots) = (500, 700, \ldots)$ for
\phone{O}.  In future work, we plan to relax this idealization (see
\cref{fn:continuous}), allowing us to investigate natural questions
such as whether \phone{u} is pronounced higher (smaller $F_1$) in
languages that also contain \phone{o} (to achieve better dispersion).

\section{Model Parameterization}

The BPP, MPP, and DPP models (\cref{sec:point-processes}) require us
to specify parameters for each vowel in ${\cal V}$.  In \cref{sec:deeppoint}, we will
accomplish this by deriving the parameters for each vowel $v_i$
from a possibly high-dimensional embedding of that vowel,
$\e(v_i) \in \Real^r$.

In \cref{sec:embed}, $\e(v_i) \in \Real^r$ will in turn be defined
as some learned function of $\f(v_i) \in \Real^k$, where
$\f : {\cal V} \mapsto \Real^k $ is the function that maps a
vowel to a $k$-vector of its measurable acoustic properties.  This
approach allows us to determine reasonable parameters even for rare
vowels, based on their measurable properties.  It will even enable us
in future to generalize to vowels that were unseen in the training set, letting
us scale to very large or infinite ${\cal V}$ (\cref{fn:continuous}).

\subsection{Deep Point Processes}\label{sec:deeppoint}

We consider deep versions of all three processes.
\paragraph{Deep Bernoulli Point Process.}
We define
  \begin{flalign}
    &\phi(v_i) = ||\e(v_i)|| \geq 0\label{eq:exponential}
  \end{flalign}

\paragraph{Deep Markov Point Process.}
The MPP employs the same unary potential as the BPP, as well as the
binary potential
\begin{align}
  &\psi(v_i, v_j) = \exp -\frac{1}{T \cdot ||\e(v_i)\!-\!
    \e(v_j)||^2} < 1 \label{eq:coulomb}
\end{align}
where the learned temperature $T > 0$ controls the relative strength
of the unary and binary potentials.

This formula is inspired by Coulomb's law for
describing the repulsion of static electrically charged
particles.  Just as the repulsive force between two particles
approaches $\infty$ as they approach each other, the 
probability of finding two vowels in the same inventory
approaches $\exp -\infty = 0$ as they approach each other.
The formula is also reminiscent of \newcite{shepard-1987}'s
``universal law of generalization,'' which says here that the probability
of responding to $v_i$ as if it were $v_j$ should fall off
exponentially with their distance in some ``psychological space''
(here, embedding space).

\paragraph{Deep Determinantal Point Process.}
For the DPP, we simply define the vector $\vec{e}_i$ to be $\e(v_i)$,
and proceed as before.

\paragraph{Summary.} In the deep BPP, the probability of a set of vowels is proportional to the product of the lengths of their embedding vectors.
The deep MPP modifies this by multiplying in {\em pairwise} repulsion terms in $(0,1)$ that increase as the vectors' endpoints move apart in Euclidean space (or as $T \rightarrow \infty$).  The deep DPP instead modifies it by multiplying in a single {\em setwise} repulsion term in $(0,1)$ that increases as the embedding vectors become more mutually orthogonal.  In the limit, then, the MPP and DPP both approach the BPP.

\subsection{Embeddings}\label{sec:embed}

Throughout this work, we simply have $\f$ extract the first $k=2$
formants, since our dataset does not provide higher formants for all
languages.\footnote{\label{fn:richf}In lieu of higher formants, we
  could have extended the vector $\f(v_i)$ to encode the binary
  distinctive features of the IPA vowel $v_i$: round, tense, long,
  nasal, creaky, etc.}For example, we have
$\f(\text{\phone{O}}) = (500, 700)$.
We now describe three possible methods for mapping $\f(v_i)$ to an
embedding $\e(v_i)$.  Each of these maps has learnable parameters.

\paragraph{Neural Embedding.}

We first consider directly embedding each vowel $v_i$ into
a vector space $\Real^r$.  We achieve this through a 
feed-forward neural net
\begin{equation}\label{eq:deep}
\e(v_i) = W_1 \tanh\left(W_0\f(v_i) +  \vec{b}_0
\right) +\vec{b}_1,
\end{equation}
\Cref{eq:deep} gives an architecture with 1 layer of nonlinearity;
in general we consider stacking $d \geq 0$ layers.  Here
$W_0 \in \Real^{r \times k}, W_1 \in \Real^{r \times
  r}, \ldots W_d \in \Real^{r \times r}$ are weight matrices, $\vec{b}_0, \ldots \vec{b}_d \in \Real^{r}$
are bias vectors, and $\tanh$ could be replaced by any pointwise
nonlinearity. 
We treat both the depth $d$ and the embedding size $r$ as hyperparameters,
and select the optimal values on a development set.

\paragraph{Interpretable Neural Embedding.}

We are interested in the special case of neural embeddings when $r=k$
since then (for any $d$) the mapping $\f(v_i) \mapsto \e(v_i)$ is a 
diffeomorphism:\footnote{Provided that our nonlinearity in \eqref{eq:deep}
is a differentiable invertible function like $\tanh$ rather than $\mathrm{relu}$.}
a smooth invertible function of $\Real^k$.  An example
of such a diffeomorphism is shown in \cref{fig:perceptual-space}.

There is a long history in cognitive psychology of mapping stimuli
into some psychological space.  The distances in this psychological
space may be predictive of generalization \cite{shepard-1987} or of
perception.  Due to the anatomy of the ear, the mapping of vowels from
acoustic space to perceptual space is often presumed to be nonlinear
\cite{rosner1994vowel,nearey2003comparison}, and there are many
perceptually-oriented phonetic scales, e.g., Bark and Mel, that carry
out such nonlinear transformations while preserving the dimensionality
$k$, as we do here.
As discussed in \cref{sec:dispersion}, vowel system typology is similarly
believed to be influenced by distances between the vowels in a latent
metric space.  We are interested in whether a constrained $k$-dimensional model of
these distances can do well in our experiments.

\paragraph{Prototype-Based Embedding.}
Unfortunately, our interpretable neural embedding is unfortunately incompatible with
the DPP.  The DPP assigns probability 0 to any vowel inventory $V$
whose $\e$ vectors are linearly dependent.  If the vectors are in
$\Real^k$, then this means that $p(V) = 0$ whenever $|V| > k$.  In our
setting, this would limit vowel inventories to size 2.\looseness=-1

Our solution to this problem is to still construct our interpretable
metric space $\Real^k$, but then map that nonlinearly 
to $\Real^r$ for some large $r$.  This latter map is
constrained.  Specifically, we choose ``prototype'' points
$\vec{\mu}_1,\ldots,\vec{\mu}_r \in \Real^k$.  These prototype points are parameters of the model: their coordinates are learned and do not necessarily correspond to any actual vowel.  We then construct $\e(v_i) \in \Real^r$ as a ``response vector'' of similarities of our vowel $v_i$ to these prototypes.  Crucially, the responses depend on
distances measured in the interpretable metric space $\Real^k$.  We use a Gaussian-density response
function, where $\vec{x}(v_i)$ denotes the representation of our vowel $v_i$ in the
interpretable space:
\begin{align}
  \e(v_i)_\ell &= w_\ell \; p(\vec{x}(v_i); \vec{\mu}_\ell, \sigma^2 I) \\
  &= w_\ell \;(2\pi \sigma^2)^{-\left(\frac{k}{2}\right)} \exp\left( \frac{-||\vec{x} - \vec{\mu}_\ell||^2}{2\sigma^2}\right). \nonumber
\end{align}
for $\ell = 1,2,\ldots,r$.  We additionally impose the constraints that
each $w_\ell \geq 0$ and $\sum_{\ell=1}^r
w_\ell = 1$. 

Notice that the sum $\sum_{\ell=1}^r \e(v_i)$ may be viewed as the density at
$\vec{x}(v_i)$ under a Gaussian mixture model.  We use this fact to
construct a prototype-based MPP as well: we redefine $\phi(v_i)$
to equal this positive density, while still defining $\psi$ via
\cref{eq:coulomb}.  The idea is that dispersion is measured in the
interpretable space $\Real^k$, and focalization is defined by
certain ``good'' regions in that space that are centered at the $r$
prototypes.

\section{Evaluation Metrics}\label{sec:evaluation}
Fundamentally, we are interested in whether our model has abstracted
the core principles of what makes a {\em good} vowel system. Our
choice of a probabilistic model provides a natural test: how surprised
is our model by held-out languages? In other words, how likely does
our model think unobserved, but attested vowel systems are?  While
this is a natural evaluation paradigm in NLP, it has not---to the best
of our knowledge---been applied to a quantitative investigation of
linguistic typology.

As a second evaluation, we introduce a {\em vowel system cloze task} that 
could also be used to evaluate non-probabilistic models. This task is defined by analogy to the traditional
semantic cloze task \cite{taylor1953cloze}, where the reader is asked
to fill in a missing word in the sentence from the context. In our
vowel system cloze task, we present a learner with a subset of the
vowels in a held-out vowel system and ask them to predict the remaining
vowels.  Consider, as a concrete example, the general American English vowel
system (excluding long vowels) $\{$\phone{i}, \phone{I}, \phone{u},
\phone{U}, \phone{E}, \phone{\ae}, \phone{O},
\phone{A}, \phone{@}$\}$. One potential cloze task would be to
predict $\{$\phone{i}, \phone{u}$\}$ given $\{$\phone{I}, \phone{U},
\phone{E}, \phone{\ae}, \phone{O}, \phone{A},
\phone{@}$\}$ and the fact that two vowels are missing from the
inventory. Within the cloze task, we report accuracy, i.e., did we
guess the missing vowel right? We consider three versions of the cloze tasks. First, we
predict {\em one} missing vowel in a setting where exactly one
vowel was deleted. Second, we predict {\em up to one} missing vowel
where a vowel {\em may} have been deleted. Third, we predict {\em up to two}
missing vowels, where one or two vowels may be deleted.

\begin{table*}
  \centering
  \begin{adjustbox}{width=2.\columnwidth}
  \begin{tabular}{l|l|lll|lll|lll} \toprule
    & \multicolumn{1}{c|}{BPP}     & \multicolumn{1}{c}{uBPP} & \multicolumn{1}{c}{uMPP} & \multicolumn{1}{c|}{uDPP}     & \multicolumn{1}{c}{iBPP} & \multicolumn{1}{c}{iMPP} & \multicolumn{1}{c|}{iDPP}     & \multicolumn{1}{c}{pBPP} & \multicolumn{1}{c}{pMPP} & \multicolumn{1}{c}{pDPP} \\ \cmidrule(r){2-2} \cmidrule(r){3-5} \cmidrule(r){6-8} \cmidrule(r){9-11}
    { x-ent }    &  8.24    & 8.28     & 8.08  & 8.00    & 13.01 & 11.50 &  \ \ \xmark & 12.83 & 10.95 & 10.29 \\
    { cloze-1}   &  69.55\% & 69.55\%  & 72.05\% & 73.18\% & 64.13\% & 67.02\% &  \ \ \xmark & 65.13\% & 68.18\% & 68.18\%  \\ 
    { cloze-01}  &  60.00\% & 60.00\%   & 61.01\% & 62.27\% & 61.78\% & 61.04\% &  \ \ \xmark & 61.02\% & 63.04\% & 63.63\% \\
    { cloze-012} &  53.18\% & 53.18\%  & 57.92\% & 58.18\% & 39.04\% & 43.02\% &  \ \ \xmark & 40.56\% & 45.01\% & 45.46\% \\
    \bottomrule
  \end{tabular}
  \end{adjustbox}
  \caption{Cross-entropy in nats (lower is better) and cloze prediction accuracy (higher is better). ``BPP'' is a simple BPP with one parameter for each of the 53 vowels in $\cal V$.  This model does artificially well by modeling an ``accidental'' feature of our data: it is able to learn not only which vowels are popular among languages, but also which IPA symbols are popular or conventional among the descriptive phoneticists who created our dataset (see \cref{fn:ipa}), something that would become irrelevant if we upgraded our task to predict actual formant vectors rather than IPA symbols (see \cref{fn:continuous}).  Our point processes, by contrast, are appropriately allowed to consider a vowel only through its formant vector.  The ``{\em u-}'' versions of the models use the uninterpretable neural embedding of the formant vector into $\mathbb{R}^r$: by taking $r$ to be large, they are still able to learn special treatment for each vowel in $\cal V$ (which is why uBPP performs identically to BPP, before being beaten by uMPP and uDPP).  The ``{\em i-}'' versions limit themselves to an interpretable neural embedding into $\mathbb{R}^k$, giving a more realistic description that does not perform as well.  The ``{\em p-}''versions lift that $\mathbb{R}^k$ embedding into $\mathbb{R}^r$ by measuring similarities to $r$ prototypes; they thereby improve on 
    the corresponding {\em i-} versions.  For each result shown, the depth $d$ of our neural network was tuned on a development set (typically $d=2$).  $r$ was also tuned when applicable (typically $r > 100$ dimensions for the {\em u-} models and $r \approx 30$ prototypes for the {\em p-} models).}
  \label{tab:cloze} 
\end{table*} 

\section{Experiments}\label{sec:experiments}

We evaluate our models using 10-fold cross-validation over the
223 languages. We report the mean performance over the 10 folds.  The
performance on each fold (``test'') was obtained by training many
models on 8 of the other 9 folds (``train''), selecting the model that
obtained the best task-specific performance on the remaining fold
(``development''), and assessing it on the test fold. Minimization
of the parameters is performed with the L-BFGS algorithm \cite{DBLP:journals/mp/LiuN89}.
As a
preprocessing step, the first two formants values $F_1$ and $F_2$ are
centered around zero and scaled down by a factor of 1000 since the
formant values themselves may be quite large.

Specifically, we use the development fold to select among the
following combinations of hyperparameters.  For neural embeddings, we
tried $r\in \{2, 10, 50, 100, 150, 200\}$. For prototype embeddings, we
took the number of components $r \in \{20, 30, 40,
50\}$.We tried network depths $d \in
\{0, 1, 2, 3\}$.  We sweep the coefficient for an $L_2$ regularizer
on the neural network parameters.

\subsection{Results and Discussion}\label{sec:results}

\Cref{fig:perceptual-space} visualizes the diffeomorphism from formant
space to metric space for one of our DPP models
(depth $d=3$ with $r=20$ prototypes).  Similar figures can be generated for all
of the interpretable models.

We report results for cross-entropy and the cloze evaluation in
\cref{tab:cloze}.\footnote{Computing cross-entropy exactly is intractable with
the MPP, so we resort to an unbiased importance sampling scheme where
we draw samples from the BPP and reweight according to the MPP
\cite{liu2015estimating}.} Under both metrics, we see that the DPP
is slightly better than the MPP; both are better than the BPP. 
This ranking holds for each of the 3 embedding schemes.  The 
embedding schemes themselves are compared in the caption.

\begin{figure}
  \includegraphics[width=\columnwidth]{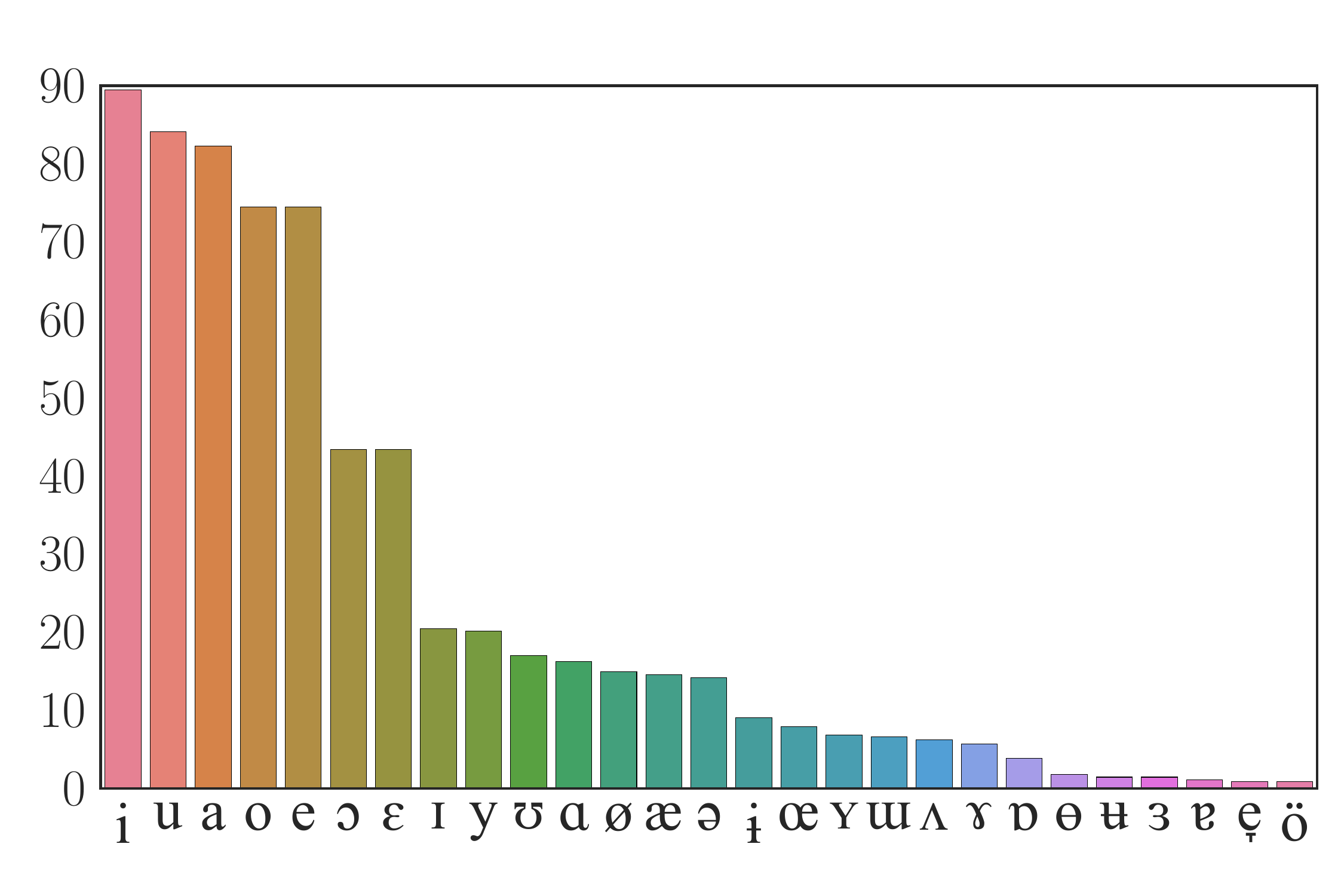}
  \caption{Percentage of the vowel inventories ($y$-axis) in the Becker-Kristal corpus \cite{becker2010acoustic} that have a given vowel (shown
    in IPA along the $x$-axis).}
  \label{fig:vowel-frequency}
\end{figure}

Within each embedding scheme, the BPP performs several points worse
on the cloze tasks, confirming that dispersion is needed to model vowel inventories
well. Still, the BPP's respectable performance shows that much of the 
structure can be capture by focalization.  As \cref{sec:point-processes}
noted, the BPP may generate well-dispersed sets, as the common vowels
tend to be dispersed already (see
\cref{fig:vowel-frequency}).  In this capacity, however, the BPP is
not explanatory as it cannot actually tell us why {\em these} vowels
should be frequent.

We mention that depth in the neural network is helpful, with
deeper embedding networks performing slightly better than depth $d=0$.

\begin{table*}
  \centering
  \begin{adjustbox}{width=2.\columnwidth}
    \begin{tabular}{lr lll lll lll} \toprule
      & \multicolumn{3}{c}{BPP} & \multicolumn{3}{c}{MPP} &  \multicolumn{3}{c}{DPP}  \\ \hdashline
    &               & \multicolumn{2}{c}{changes from $n-1$} &        & \multicolumn{2}{c}{changes from $n-1$}  &               &  \multicolumn{2}{c}{changes from $n-1$} \\
  $n$ & MAP inventory & additions  & deletions & MAP inventory & additions  & deletions & MAP inventory & additions  & deletions    \\ \cmidrule(r){1-1} \cmidrule(r){2-4} \cmidrule(r){5-7} \cmidrule(r){8-10}
 1   &   i                     & i        &  & \textschwa              & \textschwa    &               & \textschwa & \textschwa  &                  \\
 2   &   i, u                  & u        &  & i, u                    & i, u          & \textschwa    & i, u       & i, u        & \textschwa       \\
 3   &   i, u, a               & a        &  & i, u, a                 & a             &               & i, u, a     & a          &    \\
 4   &   i, u, a, o            & o        &  & i, u, a, e              & e             &               & i, u, a, o  & o          &   \\
 5   &   i, u, a, o, e         & e        &  & i, u, a, e, \textschwa  & \textschwa    &               & i, u, a, o, \textschwa   &  o \\  \bottomrule
\end{tabular}   
\end{adjustbox}
  \caption{Highest-probability inventory of each size according to our three models (prototype-based
    embeddings and $d=3$). The MAP configuration is computed by brute-force enumeration
  for small $n$. }
\label{tab:map}
\end{table*}

Finally, we identified each model's favorite complete vowel system of size $n$ (\cref{tab:map}). For the BPP, this is simply the $n$ most probable vowels.  Decoding the DPP and MPP is NP-hard, but we found the best system by brute force (for small $n$).  The dispersion in these models predicts different systems than the BPP.

\saveforCR{\paragraph{Visualization of Perceptual Space.}}

\section{Discussion: Probabilistic Typology}\label{sec:related-work}

\paragraph{Typology as Density Estimation?}
Our goal is to define a {\em universal} distribution over all possible vowel inventories.  Is this appropriate?
We regard this as a natural approach to typology, because it directly describes which kinds of linguistic systems are more or less common.  Traditional implicational universals (``all languages with $v_i$ have $v_j$'') are softened, in our approach, into conditional probabilities such as ``$p(v_j \in V \mid v_i \in V) \approx 0.9$.'' Here the 0.9 is not merely an empirical ratio, but a smoothed probability derived from the complete estimated distribution.  It is meant to make predictions about unseen languages.

Whether human language learners exploit any properties of this distribution\footnote{This could happen because learners have evolved to expect the languages (the Baldwin effect), or because the languages have evolved to be easily learned (universal grammar).}is a separate question that goes beyond typology.  \newcite{jakobson1941kindersprache} did find that children acquired phoneme inventories in an order that reflected principles similar to dispersion (``maximum contrast'') and focalization.

At any rate, we estimate the distribution given some set of attested systems that are assumed to have been drawn IID from it.  One might object that this IID assumption ignores evolutionary relationships among the attested systems, causing our estimated distribution to favor systems that are coincidentally frequent among current human languages, rather than being natural in some timeless sense.  We reply that our approach is then appropriate when the goal of typology is to estimate the distribution of {\em actual} human languages---a distribution that can be utilized in principle (and also in practice, as we show) to predict properties of {\em actual} languages from outside the training set.  

A different possible goal of typology is a theory of {\em natural} human languages.  This goal would require a more complex approach.  One should not imagine that natural languages are drawn in a vacuum from some single, stationary distribution.  Rather, each language is drawn {\em conditionally} on its parent language.  Thus, one should estimate a stochastic model of the evolution of linguistic systems through time, and identify ``naturalness'' with the directions in which this system tends to evolve.

\paragraph{Energy Minimization Approaches.}
The traditional energy-based approach \cite{liljencrants1972numerical} to vowel simulation
minimizes the following objective (written in our notation):
\begin{equation}
  {\cal E}(m) = \sum_{1 \leq i < j \leq m} \frac{1}{||\eenergy(v_i) - \eenergy(v_j)||^2},
  \label{eq:energy}
\end{equation}
where the vectors $\eenergy(v_i) \in \Real^r$ are not spit out of a deep network,
as in our case, but rather directly optimized.  
\newcite{liljencrants1972numerical} 
propose a coordinate descent algorithm to optimize ${\cal E}(m)$.  While this
is not in itself a probabilistic model, they generate diverse vowel systems through 
random restarts that find different local optima (a kind of {\em deterministic} evolutionary mechanism).  
We note that \cref{eq:energy} assumes that the number of vowels $m$ is given, 
and only encodes a notion of dispersion.  \newcite{roark2001explaining} subsequently extended \cref{eq:energy}
to include the notion of focalization. 

\begin{figure}
  \includegraphics[width=\columnwidth]{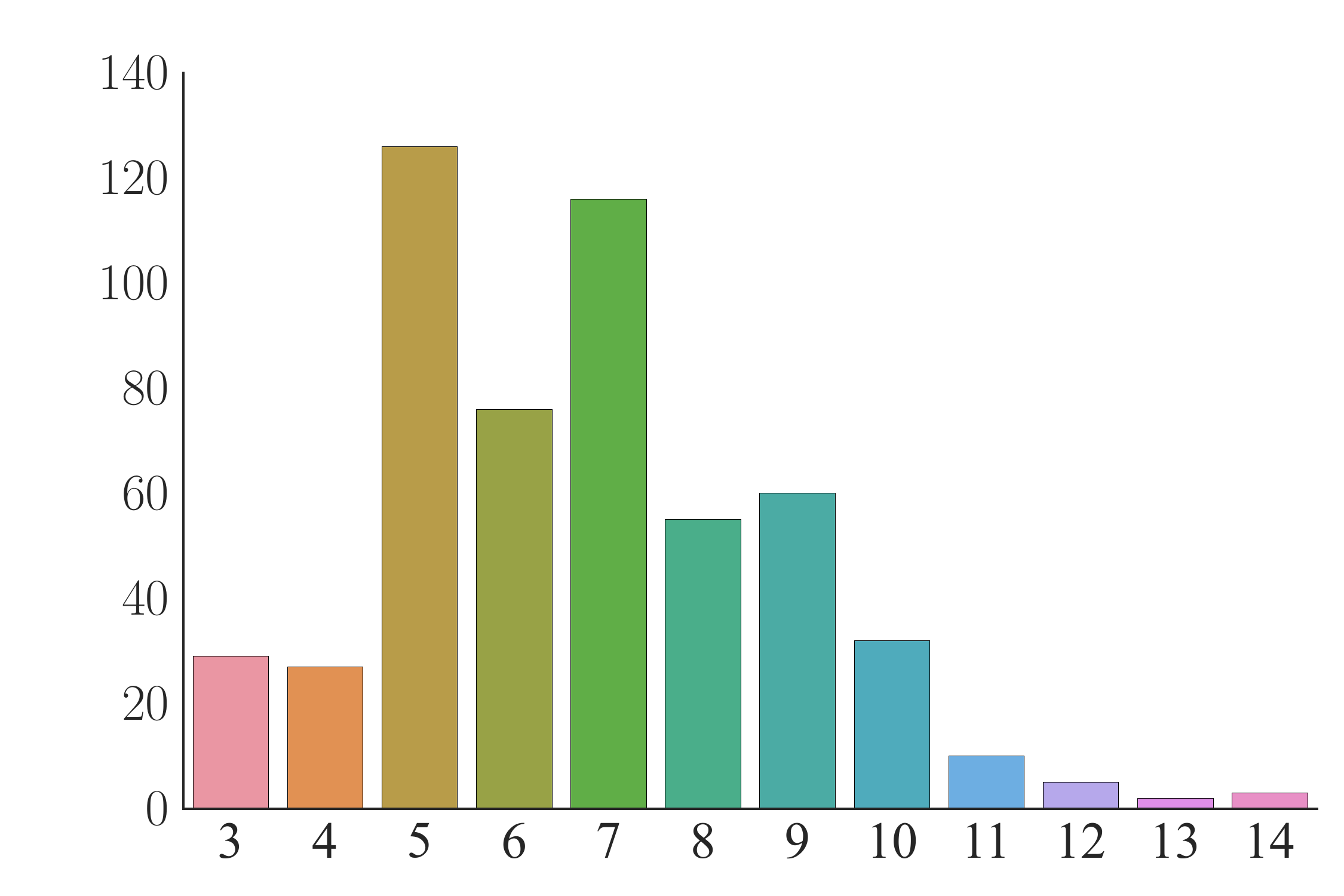}
  \caption{Histogram of the sizes of different vowel inventories in
    the corpus. The $x$-axis is the size of the vowel inventory and
    the $y$-axis is the number of inventories with that
    size.}
  \label{fig:vowel-inventories}
\end{figure}

\paragraph{Vowel Inventory Size.}
A fatal flaw of the traditional energy minimization paradigm is that it has no clear way to
compare vowel inventories of different sizes. The problem is quite
crippling since, in general, inventories with fewer vowels will have
lower energy. This does not match reality---the empirical distribution
over inventory sizes (shown in \cref{fig:vowel-inventories}) shows that the mode is
actually 5 and small inventories are uncommon: no 1-vowel
inventory is attested and only one 2-vowel inventory is known.
A probabilistic model over {\em all} vowel systems must
implicitly model the size of the system. Indeed, our models pit all potential inventories against each
other, bestowing the extra burden to match the
empirical distribution over size.

\paragraph{Frequency of Inventories.}
Another problemis the inability to model frequency. While for
inventories of a modest size (3-5 vowels) there are very few unique
attested systems, there is a plethora of attested larger vowel
systems.  The energy minimization paradigm has no principled manner to
tell the scientist how likely a novel system may be. Appealing
again to the empirical distribution over attested vowel systems, we
consider the relative diversity of systems of each size. We graph
this in \cref{fig:vowel-inventories}. Consider all vowel systems of
size 7. There are ${ |{\cal V}| \choose 7 }$ potential inventories,
yet the empirical distribution is remarkably peaked.Our probabilistic models have the advantage in this context as well, as they naturally
quantify the likelihood of an individual inventory.

\paragraph{Typology is a Small-Data Problem.}
In contrast to many common problems in applied NLP, e.g.,
part-of-speech tagging, parsing and machine translation, the modeling
of linguistic typology is fundamentally a ``small-data'' problem. Out of
the 7105 languages on earth, we only have linguistic annotation for
2600 of them \cite{wals-s1}.  Moreover, we only have
{\em phonetic and phonological annotation} for a much smaller set of
languages---between 300-500 \cite{wals-2}. Given the paucity of data,
overfitting on only those attested languages is a dangerous
possibility---just because a certain inventory has never been
attested, it is probably wrong to conclude that it is impossible---or
even improbable---on that basis alone.  By analogy to language
modeling, almost all sentences observed in practice are novel
with respect to the training data, but we still must employ a
principled manner to discriminate high-probability sentences (which
are syntactically and semantically coherent) from low-probability
ones.  Probabilistic modeling provides a natural paradigm for
this sort of investigation---machine learning has developed
well-understood smoothing techniques, e.g., regularization with tuning
on a held-out dev set, to avoid overfitting in a small-data scenario.

\paragraph{Related Work in NLP.}
Various point processes have been previously applied to potpourri of
tasks in NLP. Determinantal point processes have found a home in the
literature in tasks that require diversity. E.g., DPPs have achieved
state-of-the-art results on multi-document document summarization
\cite{KuleszaT11}, news article selection \cite{AffandiKF12}
recommender systems \cite{GartrellPK16}, joint clustering of verbal
lexical semantic properties \cite{reichart2013improved}, {\em inter alia}.
Poisson point processes have also been applied to NLP
problems: \newcite{yee2015modeling} model the emerging topic on social
media using a homogeneous point process and \newcite{LukasikCB15} apply
a log-Gaussian point process, a variant of the Poisson
point process, to rumor detection in Twitter.
We are unaware of previous attempts to probabilistically model vowel inventory typology.

\paragraph{Future Work.}
This work lends itself to several technical extensions.  One could expand the function $\f$ to more completely characterize each vowel's acoustic properties, perceptual properties, or distinctive features (\cref{fn:richf}).  One could generalize our point process models to sample finite subsets from the {\em continuous} space of vowels (\cref{fn:continuous}).  One could consider augmenting the MPP with a new factor that explicitly controls the size of the vowel inventory.  Richer families of point processes might also be worth exploring.  For example, perhaps the vowel inventory is generated by some temporal mechanism with latent intermediate steps, such as sequential selection of the vowels or evolutionary drift of the inventory.  Another possibility is that vowel systems tend to reuse distinctive features or even follow factorial designs, so that an inventory with creaky front vowels also tends to have creaky back vowels.

\section{Conclusions}

We have presented a series of point process models for the modeling of
vowel system inventory typology with the goal of a mathematical
grounding for research in phonological typology. All models were
additionally given a deep parameterization to learn representations
similar to perceptual space in cognitive science.  Also, we motivated
our preference for {\em probabilistic} modeling in linguistic typology
over previously proposed computational approaches and argued it is a
more natural research paradigm. Additionally, we have introduced
several novel evaluation metrics for research in vowel-system
typology, which we hope will spark further interest in the area. Their
performance was empirically validated on the
Becker-Kristal corpus, which includes data from over 200
languages.

\section*{Acknowledgments}
The first author was funded by an NDSEG
graduate fellowship, and the second author by NSF grant IIS-1423276.
We would like to thank Tim Vieira and Huda Khayrallah for
helpful initial feedback.

\bibliographystyle{acl_natbib}

\bibliography{732}

\end{document}